\documentclass{amia}
\usepackage{diagbox}
\usepackage{lipsum} 
\usepackage{bm}
\usepackage{times}
\usepackage{helvet}
\usepackage{courier}
\usepackage{times}
\usepackage{latexsym}
\usepackage{amsmath}
\usepackage{algorithm}
\usepackage{algpseudocode}
\usepackage{url}
\usepackage{multirow}
\usepackage{enumitem}
\usepackage{array}
\usepackage{csquotes}
\usepackage{booktabs}
\usepackage[symbol]{footmisc}
\usepackage{threeparttable}
\usepackage{graphicx}
\usepackage{dblfloatfix} 
\usepackage{caption}
\usepackage[labelformat=simple]{subcaption}

\usepackage{adjustbox}
\usepackage{multirow}
\usepackage{hyperref}
\usepackage{xcolor}
\usepackage{float}
\usepackage{longtable}
\usepackage {tcolorbox}
\usepackage{graphicx}
\usepackage{amsmath}
\usepackage{amssymb}
\usepackage{subcaption}
\usepackage{tabularx}
\usepackage{makecell}
\usepackage{mathtools}

\newcolumntype{P}[1]{>{\centering\arraybackslash}p{#1}}

\begin{document}

\title{BioInstruct: Instruction Tuning of Large Language Models for Biomedical Natural Language Processing}
\author{Hieu Tran, BSc$^1$, Zhichao Yang, MSc$^1$, Zonghai Yao, MSc$^1$, Hong Yu, PhD$^{1, 2, 3, 4}$}
\institutes{
    $^1$ Manning College of Information and Computer Sciences, University of Massachusetts Amherst, MA, USA\\
    $^2$ Department of Medicine, University of Massachusetts Medical School, Worcester, MA, USA\\
    $^3$ Center for Biomedical and Health Research in Data Sciences, Miner School of Computer and Information Sciences, University of Massachusetts Lowell, MA, USA\\
    $^4$ Center for Healthcare Organization and Implementation Research, VA Bedford Health Care, MA, USA\\
}
\maketitle
\textbf{Corresponding author:} Hong Yu, PhD

Center for Biomedical and Health Research in Data Sciences, Miner School of Computer and Information Sciences, University of Massachusetts Lowell, MA, USA 

Phone: 1 978-934-3620  
Email: Hong\_Yu@uml.edu

\section*{Abstract}
\textbf{Objective:} 
To enhance the performance of large language models (LLMs) in biomedical natural language processing (BioNLP) by introducing a domain-specific instruction dataset and examining its impact when combined with multi-task learning principles.

\textbf{Materials and Methods:} 
We created the  \textit{BioInstruct}, comprising 25,005 instructions to instruction-tune LLMs(LLaMA 1 \& 2, 7B \& 13B version). The instructions were created by prompting the GPT-4 language model with three-seed samples randomly drawn from an 80 human curated instructions. We employed Low-Rank Adaptation(LoRA) for parameter-efficient fine-tuning. We then evaluated these instruction-tuned LLMs on several BioNLP tasks, which can be grouped into three major categories: question answering(QA), information extraction(IE), and text generation(GEN). We also examined whether categories(e.g., QA, IE, and generation) of instructions impact model performance.

\textbf{Results and Discussion:} 
Comparing with LLMs without instruction-tuned, our instruction-tuned LLMs demonstrated marked performance gains: 17.3\% in QA on average accuracy metric, 5.7\% in IE on average F1 metric, and 96\% in Generation tasks on average GPT4 score metric. Our 7B-parameter instruction-tuned LLaMA 1 model was competitive or even surpassed other LLMs in the biomedical domain that were also fine-tuned from LLaMA 1 with vast domain-specific data or a variety of tasks. Our results also show that the performance gain is significantly higher when instruction fine-tuning is conducted with closely related tasks. Our findings align with the observations of multi-task learning, suggesting the synergies between two tasks.

\textbf{Conclusion:}
The \textit{BioInstruct} dataset serves as a valuable resource and instruction tuned LLMs lead to the best performing BioNLP applications. 


\textbf{Keywords:} Instruction Tuning, Large Language Models, Question Answering, Natural Language Inference, Information Extraction, Text Generation, Multi-task Learning

\textbf{Word count: 3992}

\section*{Introduction}
LLMs, including GPTs, have made significant impact on natural language processing (NLP) applications~\cite{brown2020language,Sanh2021,chowdhery2022palm,longpre2023flan,openai2023gpt4,Yang2023.10.26.23297629}. In the clinical domain, efforts have been made to fine-tune LLMs \cite{alsentzer2019publicly, peng2019transfer, van2021clinical, lehman2023we}. Yet, this method can be resource-intensive and is at risk of overfitting, more so when faced with limited or low-quality clinical data \cite{alsentzer2019publicly}. In contrast, "instruction tuning" emerges within the NLP community as a promising alternative to such exhaustive fine-tuning of LLMs \cite{wei2021finetuned}. Stemming from Instruction Fine-tuning \cite{chung2022scaling}, it enables models to adapt to and perform new tasks more effectively through natural language instructions alone. Innovations by Mishra et al. \cite{mishra-etal-2022-cross} and Wang et al. \cite{wang2022super} have laid the groundwork for instruction tuning by harmonizing crowdsourced instructions. This approach, expanded upon by Sanh et al. \cite{Sanh2021} and Wei et al. \cite{wei2021finetuned}, strives for adaptability to novel instructional tasks. Subsequent efforts, like those by Chung et al. \cite{chung2022scaling}, amplify the technique, spotlighting task diversity, augmented model scale, and integrated chain-of-thought, with Ouyang et al. \cite{ouyang2022training} introducing a unique reinforcement learning perspective. Nevertheless, despite its notable progress in general NLP scenarios, the biomedical field finds itself underrepresented, primarily attributed to the missing tailored instruction sets \cite{rasmy2021med, alsentzer2019publicly}. Addressing this lacuna, our study introduces a comprehensive BioNLP instruction dataset, curated with limited human intervention.

Specifically, we introduce \textbf{BioInstruct}, a dataset comprising more than 25,000 natural language instructions along with their corresponding inputs and outputs. Drawing inspiration from recent work that leverages the GPT language model for data generation \cite{wang2022self}, we collect BioInstruct in a fully automated manner. We prompt a pre-trained LLM, GPT-4, with a sample of three examples (as seeds) from our manually collected triplets (instructions, input, output), then instruct the model to generate new instructions as illustrated in Appendix Table \ref{fig:prompt}. Through this process, we automatically produce over 25,000 diverse triplets consisting of instructions, inputs, and outputs spanning a range of biomedical NLP tasks. We subsequently use BioInstruct to fine-tune both LLaMA 1 \cite{touvron2023LLaMA} and LLaMA 2 \cite{touvron2023LLaMA2}.

To evaluate LLM's ability for the biomedical applications, we introduced a benchmark including the several BioNLP tasks which can be grouped into three major categories: question answering, information extraction, and text generation.

\begin{itemize}
    \item \textbf{QA} (Question Answering) category include Question Answering and Natural Language Inference (NLI) tasks play a pivotal role in healthcare by allowing accurate retrieval of specific knowledge from vast medical repositories, aiding doctors in diagnostics and treatment planning \cite{jin2022biomedical}.
    
    \item \textbf{IE} (Information Extraction) category aims to identify and extract relevant data from unstructured clinical text automatically, enabling healthcare professionals to make well-informed decisions, thus improving patient care \cite{wang2018clinical}.
    
    \item \textbf{Text Generation} category has the potential to revolutionize patient care by summarizing conversation into clinical notes \cite{ben-abacha-etal-2023-overview,Wang2023NoteChatAD, wang2023umass_bionlp,krishna2020generating,yao2023improving} or generating patient's assessment given patients symptom \cite{zeng-etal-2020-meddialog, yang-yu-2020-generating}.
\end{itemize}

Our findings offer insights into the diverse impacts of BioInstruct across these tasks. 
We find that LLMs fine-tuned with BioInstruct outperformed LLMs without BioInstruct in QA, IE, Generation tasks by 17.3\%, 5.7\%, 96\% respectively. Driven by the relatively low improvements in IE, we conducted data ablation experiments to explore the type of tasks in BioInstruct that contribute to these improvements. We observe that contributing fine-tuning tasks is unique to each evaluation task. When evaluating text generation, LLMs fine-tuned on all tasks from BioInstruct outperforms LLMs fine-tuned on any single task. But this is not the case when evaluating on QA and IE.

In addition, experiments show that fine-tuning a 7B-parameter LLaMA 1 model \cite{touvron2023LLaMA} on BioInstruct can perform competitively or even outperform baseline models such as PMC-LLaMA 7B \cite{wu2023pmc}, Asclepius 7B \cite{kweon2023publicly}, MedAlpaca 7B \cite{han2023medalpaca} and ChatDoctor \cite{yunxiang2023chatdoctor} across several benchmarks. Notably, all these models, including ours, are derived from the foundational LLaMA 1 model \cite{touvron2023LLaMA}, ensuring a fair basis for comparison. Specifically, we note that our instructed LLaMA 1 model outperforms in 4 out of 4 question-answering tasks when compared with Asclepius \cite{kweon2023publicly}, PMC-LLaMA \cite{wu2023pmc} and it also exceeds MedAlpaca \cite{han2023medalpaca} in natural language inference task. This suggests that BioInstruct is particularly effective for augmenting the model to apply medical knowledge. 
We also notice a positive correlation between the number of generated examples and downstream task performance, implying that the performance of models trained on BioInstruct can be further enhanced simply by expanding its size. To summarize, our research contributions are as follows:

\begin{itemize}
[leftmargin=.2in,topsep=2pt]
\setlength\itemsep{0.01em}
\vspace{-0.2em}
    \item We introduce a new benchmark to evaluate LLMs ability on three categories of BioNLP tasks: QA to evaluate medical knowledge, IE to evaluate clinical extraction, and text generation to evaluate applied clinical skills.
    
    \item We introduce a new Instruction Tuning data BioInstruct, which specifically tailored for the biomedical domain. We find that LLMs fine-tuned on BioInstruct significantly improve performance on the benchmark compared to competitive baselines.
    
    \item We further explore the type of tasks in BioInstruct that contribute to these improvements through the Multi-Task Learning framework. We observe that contributing to fine-tuning tasks is dependent on each evaluation task. This inspires future work to predict contributing tasks given a new evaluation task.
\end{itemize}

\section*{Related Work}

\paragraph{LLMs in BioNLP} The use of LLMs in the field of natural language processing (NLP) has shown remarkable potential and achieved significant milestones. The fine-tuning of these models on specific tasks has resulted in breakthroughs across a wide range of applications, including translation, text generation, and question-answering. In BioNLP, LLMs have also begun to play a crucial role. These models, when fine-tuned on biomedical text corpora, have shown promising results in medical information extraction, biomedical literature summarization, and answering medical questions. Models like BioBERT \cite{lee2020biobert}, ClinicalBERT \cite{alsentzer2019publicly}, and the recent MedAlpaca \cite{han2023medalpaca} which introduces an open-source collection of medical conversational AI models and training data tailored for LLMs, have advanced biomedical text mining capabilities. Additionally, fine-tuned LLMs such as PMC-LLaMA \cite{wu2023pmc}, Asclepius \cite{kweon2023publicly} and ChatDoctor \cite{yunxiang2023chatdoctor} have shown impressive performance in various biomedical tasks, setting new benchmarks in the field.

\paragraph{Traditional Fine-Tuning vs Instruction Tuning}
Traditional fine-tuning and instruction tuning offer contrasting methods for adapting large language models (LLMs) to specific tasks. Traditional fine-tuning, as used in models like BioBERT \cite{lee2020biobert} and ClinicalBERT \cite{alsentzer2019publicly}, involves extensive retraining of a pre-trained model on task-specific datasets. This approach often yields highly specialized models but requires substantial, high-quality data and can be prone to overfitting. On the other hand, instruction tuning, exemplified by works like \cite{wei2021finetuned}, focuses on the model's ability to follow natural language instructions, promoting flexibility and generalizability across various tasks with less dependency on extensive task-specific data. This approach is particularly beneficial in scenarios where acquiring large annotated datasets is challenging. However, instruction tuning may not always achieve the same level of task-specific accuracy as traditional fine-tuning with a dedicated dataset. The choice between these methodologies depends on factors like data availability, adaptability needs, and desired task performance.

\paragraph{Instruction Tuning in BioNLP}
Obtaining large-scale supervised data can be expensive and time-consuming. A popular solution in recent literature has been to use language models (LMs) for automatic data generation and augmentation \cite{schick-schutze-2021-generating, meng2023tuning}. Different from that, Self-Instruct \cite{wang2022self} is not specific to a particular task (say, QA or NLI) but to bootstrap new task definitions that may not have been defined before by NLP practitioners (though potentially still important for real users) \cite{honovich2022unnatural, wang2022super, brown2020language}.
Recent work by \cite{wei2021finetuned} highlighted that instruction tuning -- fine-tuning language models on a collection of datasets described via instructions substantially improves zero-shot performance of LLMs on unseen tasks. 
In parallel with our work, Zhang et al. \cite{zhang2023alpacare} also propose generating instruction data with GPT-4 models and fine-tuning with the LLaMA model (so called AlpaCare). The major differences are that 1) they use a set of clinician-crafted tasks as their seed tasks, resulting in a different distribution of generated tasks; 2) they evaluate the performance of their instructed model on different benchmarks and employ different metrics.

\section*{Methodology}
Our methodology incorporates two primary components: the creation of our BioInstruct dataset and the subsequent fine-tuning of several LLMs.

\begin{figure*}[ht!]
\centering

\begin{subfigure}{0.39\textwidth}
\centering
\includegraphics[width=\textwidth]{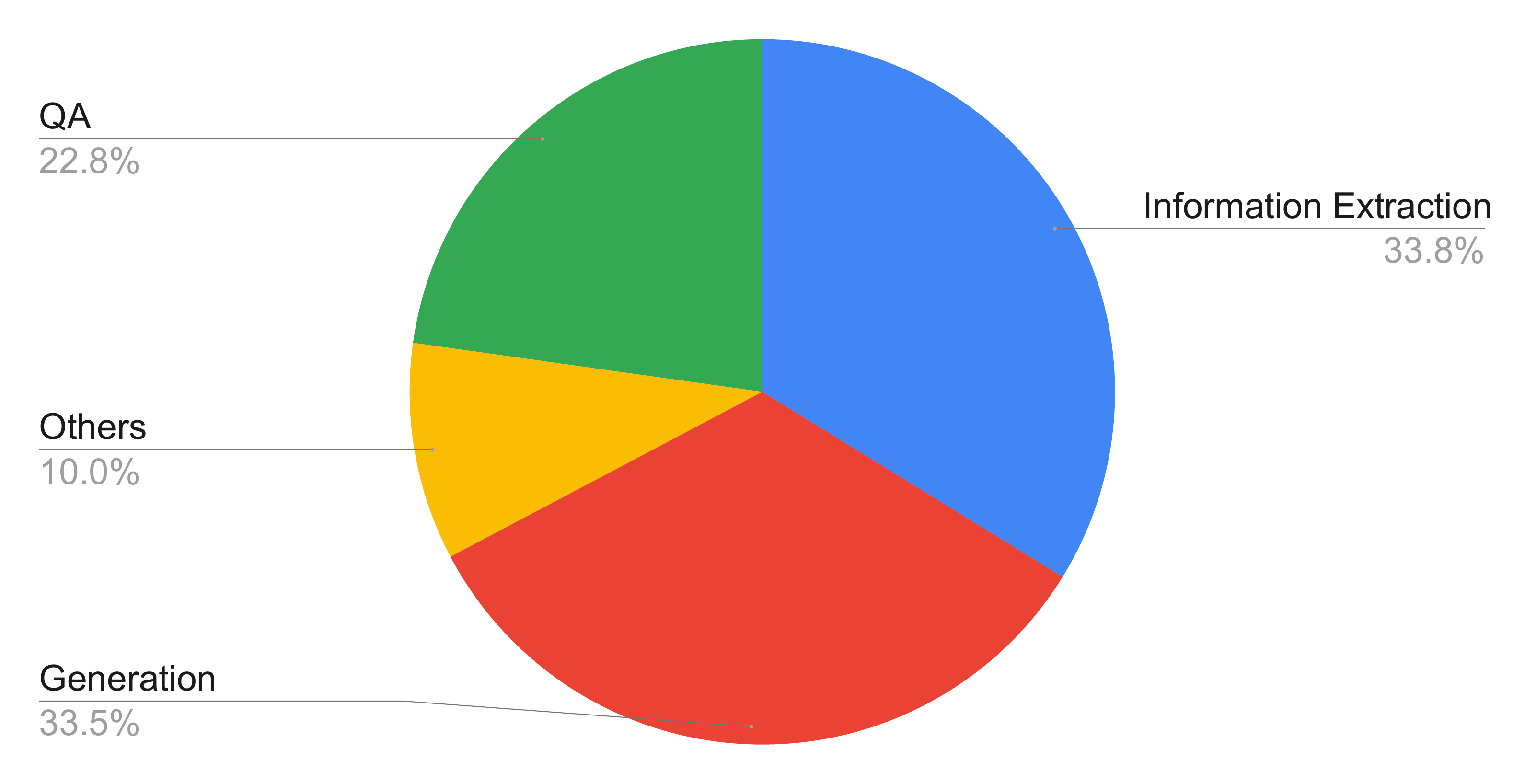}
\caption{Task type distribution of 25,005 natural language instructions}
\label{subfig:tasktype}
\end{subfigure}
\hfill
\begin{subfigure}{0.6\textwidth}
\centering
\includegraphics[width=\textwidth]{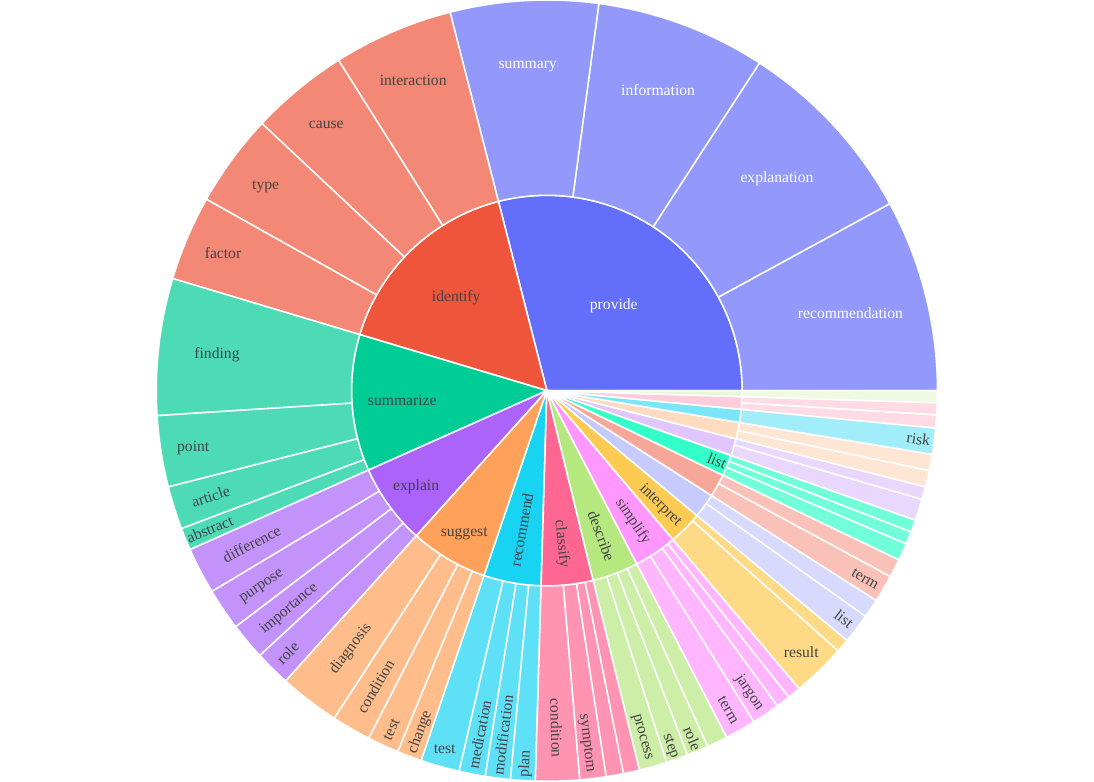}
\caption{The top 20 most common root verbs (inner circle) and their top 4 direct noun objects (outer circle) in the generated instructions.}
\label{subfig:overview}
\end{subfigure}

\caption{Distribution of our BioInstruct dataset}
\label{fig:combined}

\end{figure*}

\subsection*{Instruction Collection}

We introduce BioInstruct, a dataset comprising 25,005 natural language instructions tailored to a broad array of biomedical and clinical natural language processing tasks. Each entry in this dataset is structured with a natural language instruction, an associated input, and the expected output resulting from the instruction's execution.

Taking inspiration from the \enquote{Self-Instruct} methodology \cite{wang2022self}, the collection of BioInstruct is a fully automated process. This process requires only an initial set of 80 manually constructed seed tasks, which can be produced within roughly three hours of human effort. These seed examples span a diverse range of biomedical and clinical NLP tasks, covering areas such as answering biomedical questions, summarizing, assessing eligibility for clinical trials, and determining differential diagnosis, among others (some examples of seed tasks are shown in Appendix Table { \ref{fig:examples}}).  During the data collection phase, we prompted the pretrained GPT-4 language model with three examples randomly selected from seed tasks, guiding it to generate new samples (illustrated in Appendix Table { \ref{fig:prompt}}). We prompted GPT-4 to create 25,005 natural language instructions in July 2023.

Among the GPT-4 created instructions, we plot the top 20 most common root verbs and their top 4 direct noun objects of BioInstruct dataset in \figurename{ \ref{fig:combined}}. Overall, we see quite diverse intents and textual formats in these instructions. 
Additionally, we used GPT-4 to classify the instructions into major categories such as Question Answering, Generation, Information Extraction, and Others(e.g. trial recommendation). We validate the GPT-4 classification performance through 1000 test examples. The accuracy is 87.3\%. Moreover, to provide a deeper insight into the reliability of these category predictions, we conducted an inter-rater reliability analysis with three independent raters who classified each example into the designated major categories. This process resulted in a Krippendorff's alpha of 0.83, reflecting a strong consensus among the raters and underscoring the consistency and precision of our categorization approach. The high degree of agreement, as indicated by this alpha value, further validates the integrity and reliability of our methodology. The distribution of these categories is visually represented in the pie chart shown in \figurename{ \ref{fig:combined}}.
The data generation prompt consists of examples in a structured format, making it easier for the pre-trained GPT-4 language model to generate.
Each example in the dataset contains three fields:

\begin{enumerate}
[leftmargin=.2in,topsep=2pt]
\setlength\itemsep{0.01em}
\vspace{-0.2em}
    \item An \textbf{instruction} describing the task (e.g. “Given a lengthy patient education material, provide a concise summary that preserves the crucial information, while ensuring it remains accessible for patients”).
    \item The \textbf{input} argument that instantiates the instruction, creating a specific example of the task.
    \item A textual \textbf{output} reflecting a correct execution of the instruction given the input arguments.

\end{enumerate}

For promoting diversity, we only incorporate a new instruction into the task pool if its ROUGE-L similarity with present instructions falls below 0.7. Instructions with particular keywords (like image, picture, graph), which are typically not handled by LMs, are disregarded.  Invalid generations are identified and filtered out based on heuristics including instruction is too long (more than 150 words) or too short (less than 3 words), instance output is a repetition of the input

\subsection*{Instruction Tuning}

Instruction Tuning refers to continually training an LLM with NLP tasks formatted as natural language instructions and model responses treated as task outputs \cite{wei2021finetuned}. Further to solving structured natural language tasks \cite{taoristanford} turned an English-centric LLM, LLaMA 1 \cite{touvron2023LLaMA}, into an open-ended chat model. It delivers GPT-like performance for English by training on distilled data from GPT itself \cite{wang2022self}. Our work follow the similar approach performing instruction tuning on LLaMA 1 \cite{touvron2023LLaMA} and LLaMA 2 \cite{touvron2023LLaMA2} using our BioInstruct dataset.

The core idea behind Instruction Tuning is to utilize natural language as a versatile interface, allowing models to process instructions in the same way they would handle any other text input. This significantly broadens their usability across various domains and tasks, leveraging their pre-trained capabilities to interpret and act upon instructions directly. Instruction Tuning thus encourages models to become more flexible in their application, moving towards an intuitive understanding of tasks through human-like instructions.

An example of how instruction tuning is performed can be seen in what we show in Appendix Table \ref{fig:instruction-tuning}. This table presents a typical instruction tuning prompt format, where an instruction clearly defines a task, followed by an input providing contextual background, and the model generates an appropriate response based on the instruction. This approach exemplifies the Instruction Tuning process, illustrating its implementation in training models to perform specialized tasks by utilizing task descriptions and inputs to guide the generation of precise outputs.

\subsection*{Multi-task Exploration}
To find out which task in BioInstruct contributed the most for the evaluation benchmark. We first grouped the samples in BioInstruct into the following BioNLP application categories: QA, IE, generation, and other tasks, each consisting of 22.8\%, 33.8\%, 33.5\%, and 10\%, respectively. 
We then fine-tuned on a subset (following single or multiple tasks) of BioInstruct tasks as mentioned previously. We then evaluated models on the benchmark. We exclude NLI as there are few samples in the BioInstruct dataset.

\section*{Experimental Setup}
In this section, we detail our experimental setup, the datasets employed, and the evaluation strategy adopted for assessing the performance of our instruction-tuned LLMs in various BioNLP tasks. A 2-sided t test was used to determine the significance of improvements between different models. All significance tests were evaluated at $\alpha=0.05$. Furthermore, all experiments were conducted using two Nvidia A100 GPUs, each with 40 GB of memory. The CPU used was an Intel Xeon Gold 6230 processor, and the system was equipped with 192 GB of RAM.

\subsection*{Benchmark Datasets}

Our experiments spanned multiple subtasks, covering major NLP tasks within the biomedical domain. We focused on recent tasks to prevent data leakage as GPT4 was trained for information up until September 2021:

\noindent \textbf{Multiple Choice Question Answering (MCQA):}
    \begin{itemize}
        [leftmargin=.2in,topsep=2pt]
        \setlength\itemsep{0.01em}
        \vspace{-0.2em}
        \item \textbf{MedQA-USMLE} \cite{jin2021disease}: A subtask containing multiple-choice questions from the United States Medical Licensing Examination.
        \item \textbf{MedMCQA} \cite{pal2022medmcqa}: Features medical multiple-choice questions source from various textbooks and clinical scenarios.
        \item \textbf{PubMedQA} \cite{jin2019pubmedqa}: A benchmark dataset for biomedical research question answering derived from PubMed abstracts.
        \item \textbf{BioASQ MCQA} \cite{tsatsaronis2015overview}: A comprehensive subtask designed for biomedical semantic indexing and question answering.
    \end{itemize}
   \textbf{Natural Language Inference (NLI)}:
    \begin{itemize}
        [leftmargin=.2in,topsep=2pt]
        \setlength\itemsep{0.01em}
        \vspace{-0.2em}
        \item \textbf{MedNLI} \cite{shivade2019mednli}: A medical NLI subta crafted from clinical narratives, which tasks models with determining the relationship between premise and hypothesis sentences. We excluded NLI for multi-task exploration, as there were few samples in the BioInstruct dataset.
    \end{itemize}
   \textbf{Clinical Information Extraction}:
    \begin{itemize}
        [leftmargin=.2in,topsep=2pt]
        \setlength\itemsep{0.01em}
        \vspace{-0.2em}
        \item \textbf{Medication Status Extraction} \cite{agrawal2022large}: A subtask focusing on the extraction of medication-related information from clinical narratives, such as drug names, dosages, and administration routes.
        \item \textbf{Clinical Coreference Resolution} \cite{agrawal2022large}: This involves identifying phrases in clinical texts that refer to the same entity, facilitating understanding patient narratives.
    \end{itemize}
    \textbf{Generation Task about clinical skills:}
    \begin{itemize}
        [leftmargin=.2in,topsep=2pt]
        \setlength\itemsep{0.01em}
        \vspace{-0.2em}
        \item \textbf{Conv2note} 
        \cite{ben-abacha-etal-2023-overview}: This is a subtask (A) of the MediQA-Chat 2023 Challenge. The goal is to generate a section of a structured clinical note based on a given patient-doctor conversation, with an emphasis on accurately capturing medical details.
        \item \textbf{ICliniq} \cite{yunxiang2023chatdoctor}: In this subtask, given a patient's detailed description of their concerns or symptoms, the aim is to provide a compassionate, reassuring, and informative medical response that addresses their worries and offers professional guidance on the next steps or actions to take. We evaluate this using a random subset of 100 samples from the ICliniq dataset.
    \end{itemize}

\subsection*{Evaluation Metrics}

For the multiple choice QA and NLI tasks, we employed accuracy as the primary metric. 

In the clinical information extraction tasks, performance was gauged using precision, recall, F1 score, and \textbf{conditional accuracy}. The latter specifically measured the correctness of medication status classification, indicating how many statuses were correctly identified. 

For the generation task, we utilized the GPT4 API to evaluate \textbf{Coherence}, \textbf{Naturalness}, and \textbf{Completeness} scores, which assess the quality and relevance of the generated contents. This decision was informed by recent studies \cite{fu2023gptscore, liu2023g}, which demonstrated GPT-4's strong alignment with human evaluation in various generative tasks. Those works have shown GPT-4's evaluations on coherence, naturalness, and completeness to closely mirror human judgments, suggesting its reliability in assessing the nuanced aspects of generated content. In the clinical NLP domain, a previous study \cite{zhang2023huatuogpt} compared GPT-4 evaluation and human evaluation on 100 patient-doctor conversations and show that GPT-4 evaluation generally aligns with human evaluation in Figures 3 and 4 of \cite{zhang2023huatuogpt}. To further refine our evaluation, we incorporated the \textbf{Bertscore\_F1} metric. This metric, derived from the embeddings of the BERT model, gauges the semantic similarity between the generated and reference texts. Another significant metric we introduced was the \textbf{Concept\_F1}, which calculates the overlap of medical entities between two documents, offering insight into the model's grasp of medical knowledge and terminology. Additionally, the \textbf{Bleurt} score, an advanced metric designed specifically for generation tasks, was employed. It determines the alignment of the model’s output with the human-written reference, both syntactically and semantically.


\begin{table*}[t!]
\begin{center}
\resizebox{0.99\textwidth}{!}{
\begin{tabular}{l|c|c|c|c|c}
\hline
Model & \textbf{MedQA-USMLE} & \textbf{MedMCQA} & \textbf{PubmedQA} & \textbf{BioASQ MCQA} & \textbf{MedNLI} \\ \hline

AlpaCare LLaMA 1 7B	& 28.7 ± 0.0722 &  33.91 ± 0.0097 &	 62.9 ± 0.2104 & 68.43 ± 2.7543  & 33.3 ± 0.3143  \\

AlpaCare LLaMA 2 7B	& 34.88 ± 0.0466 & 36.04 ± 0.0643 & 55.52 ± 1.0427 & 67.14 ± 2.8654 & 35.96 ± 0.1209  \\

Asclepius 7B & 29.08 ± 0.2405 & 30.46 ± 0.0953 & 40.14 ± 0.3667 & 72.14 ± 1.439 & 38.83 ± 0.0302  \\

ChatDoctor & 30.92 ± 0.1097 & 31.01 ± 0.1182 &  63.08 ± 0.4116 &  74.86 ± 0.601  &  40.14 ± 1.5672  \\

MedAlpaca  7B &	37.61 ± 0.0467 & 34.37 ± 0.079	&  48.82 ± 0.3415 &	61.57 ± 5.9762  & 36.6 ± 2.1421 \\

PMC-LLaMA 7B & 27.73 ± 0.0743 &  26.94 ± 0.1485 &  55.8 ± 1.4445 &  61.43 ± 1.4289  & 33.71 ± 1.1738  \\

LLaMA 1 7B & 27.47 ± 0.0628 & 25.06 ± 0.1601 & 47.54 ± 0.5771 & 70.14 ± 2.7543  & 33.59 ± 0.9334  \\

LLaMA 1 7B Instruct & 31.5 ± 0.0038 & 31.46 ± 0.0561 & 64.36 ± 1.6479 & 81.14 ± 1.8933 & \textbf{43.27} ± 1.9985 \\

LLaMA 1 13B & 33.85 ± 0.8164 &  31.76 ± 0.0473& 54.88 ± 0.4572 & 69.86 ± 1.4991  & 34.62 ± 2.0661  \\

LLaMA 1 13B Instruct & 36.76 ± 0.121 &  34.03 ± 0.4909 & 58.75 ± 1.3148 & 84.29 ± 2.0862 & 34.63 ± 0.5435 \\

LLaMA 2 7B & 30.37 ± 1.0608 & 29.41 ± 0.0377 &  57.2 ± 0.7538 &  65.86 ± 3.1138  & 33.35 ± 0.016  \\

LLaMA 2 7B Instruct & 37.25 ± 0.3621 &  36.14 ± 0.0254 & 63.97 ± 0.7363 &  84.76 ± 4.4353 & 34.96 ± 1.3937 \\

LLaMA 2 13B & 34.89 ± 0.5802 &  32.37 ± 0.3702 &  67.76 ± 0.7174 & 73.14 ± 2.97  &  35.94 ± 1.006  \\

LLaMA 2 13B Instruct & \textbf{39.32} ± 0.0998 &  \textbf{36.54} ± 0.27 &  \textbf{71.73} ± 1.0781 &  \textbf{87.38} ± 4.8823 &  41.7 ± 0.1315 \\
\hline
\end{tabular}
}
\end{center}
\caption{\label{tab:medical-model-comparison} Zero-shot performance of the original, instructed, and some baseline models on multiple-choice QA tasks and natural language inferences tasks, $p<0.05$; 95\% confidence interval.
}
\end{table*}

\subsection*{Baseline Models}
In our study, we compared the performance of BioInstruct with several prominent baselines in the BioNLP domain:

\textbf{Asclepius} \cite{kweon2023publicly} is a specialized clinical LLM, further fine-tuned from LLaMA 1 on synthetic clinical notes derived from publicly available biomedical literature. Preliminary evaluations show that Asclepius can effectively handle tasks on real clinical notes, showing its promise for real-world healthcare applications.

\textbf{ChatDoctor} \cite{yunxiang2023chatdoctor} is a language model aiming for health assistants, that is designed to provide users with medical information, advice, and guidance. For training, it is fine-tuned from LLaMA 1 7B model with the dialogue-based instruction tuning data.

\textbf{MedAlpaca} \cite{han2023medalpaca} is a model that has been further fine-tuned on Alpaca \cite{taoristanford}, which is an instruction-tuned variant of LLaMA. Its primary focus is on assisting with medical dialogues and handling question-answering tasks.

\textbf{PMC-LLaMA} \cite{wu2023pmc} is a domain-adapted LLaMA 1 model that was pretrained on 4.8M biomedical academic papers and 30K medical textbooks, and was further fine-tuned with a comprehensive dataset tailored for instruction tuning, covering areas like medical question-answering and conversational dialogues.

\section*{Results}

\begin{table}[t!]
\begin{center}
\resizebox{\textwidth}{!}{
\begin{tabular}{l|c|c|c|c|c|c|c}
\hline
\textbf{Models} & \multicolumn{4}{c|}{\textbf{Medication Status Extraction}} & \multicolumn{3}{c}{\textbf{Coreference Resolution}} \\ \hline
 & \textbf{Precision} & \textbf{Recall} & \textbf{F1} & \textbf{Conditional ACC} & \textbf{Precision} & \textbf{Recall} & \textbf{F1} \\ \hline

PMC-LLaMA 7B & 68.08 & 78 & 67.28 ± 0.3213 & 71.21 ± 0.3311 & 56.56 & 47.37 & 45.97 ± 0.1935 \\

ChatDoctor & 63.07 & 83.56 & 67.6 ± 0.0232 & 80.31 ± 0.3565 & 58.29 & 56.18 & 54.02 ± 0.0366 \\

LLaMA 1 7B & 67.61 & 71.95 & 66.58 ± 0.0601 & 78.45 ± 0.2048 & 60.14 & 39.19 & 43.8 ± 0.1914 \\

LLaMA 1 7B Instruct & 62.83 & 84.43 & 68.97 ± 0.241 & \textbf{82.52} ± 0.6844 & 67.58 &52.52 & 55.14 ± 0.1765 \\

LLaMA 2 7B & 65.58 & \textbf{89.16} & 71.72 ± 0.2313 & 76.69 ± 0.1564 & 61.48 & 45.97 & 50.03 ± 0.3343 \\

LLaMA 2 7B Instruct & \textbf{75.44} & 82.83 & \textbf{75.63} ± 0.3701 & 82.3 ± 0.0664 & \textbf{72.58} & \textbf{58.69}& \textbf{61.24} ± 0.439 \\
\hline
\end{tabular}
}
\end{center}
\caption{\label{tab:clinical-extraction} One-shot performance of the baseline models, original LLaMA, and instructed LLaMA on the medication status extraction and coreference resolution tasks. \textbf{Conditional ACC} measured the correctness of medication status classification, indicating how many statuses were correctly identified conditional on the extracted medications, $p<0.05$; 95\% confidence interval.}
\end{table}

\begin{table*}[h!]
    \centering
    \scalebox{0.9}{
    \begin{tabular}{lcccccc}
        \hline
        Models & \multicolumn{3}{c}{GPT4 Metrics} & \multicolumn{3}{c}{Other Metrics} \\
        & Coherence & Completeness & Naturalness & Concept\_F1 & BertScore\_F1 & Bleurt \\
        \hline
        \multicolumn{7}{c}{Conv2note} \\
        
PMC-LLaMA 7B & 2.84 ± 0.0031 & 2.91 ± 0.0010 & 2.46 ± 0.0016 & 17.91 ± 0.3291 & 85.91 ± 0.5463 & 46.63 ± 0.2789 \\

ChatDoctor & 4.47 ± 0.0004 & 4.46 ± 0.0016 & 4.37 ± 0.0004 & 29.42 ± 0.0491 & 86.99 ± 0.3024 & 51.34 ± 0.1637 \\
        
LLaMA 1 7B & 2.74 ± 0.0020 & 2.76 ± 0.0016 & 2.44 ± 0.0038 & 17.87 ± 0.5227 & 85.28 ± 0.3021 & 46.95 ± 1.8605 \\

LLaMA 1 7B Instruct & \textbf{4.49} ± 0.0011  & \textbf{4.54} ± 0.0012 & \textbf{4.42} ± 0.0014 & \textbf{31.13} ± 0.1979 & \textbf{88.45} ± 0.0218 & \textbf{53.19} ± 0.0231 \\
        
LLaMA 2 7B & 2.80 ± 0.0037 & 2.96 ± 0.0009 & 2.57 ± 0.0060 & 18.93 ± 0.9404 & 84.92 ± 0.0894 & 45.95 ± 0.2743 \\

LLaMA 2 7B Instruct & 4.45 ± 0.0025 & 4.41 ± 0.0011 & 4.28 ± 0.0043 & 27.87 ± 0.1094 &88.06 ± 0.0254 & 51.67 ± 0.0252 \\
\hline
\multicolumn{7}{c}{Doctor-Patient QA} \\

PMC-LLaMA 7B & 2.31 ± 0.0060 & 2.64 ± 0.0132 & 2.34 ± 0.0039 & 15.85 ± 0.1743 & 
84.02 ± 1.6380 & 46.55 ± 0.1897 \\

ChatDoctor & 4.81 ± 0.0002 & 4.86 ± 0.0007 & 4.62 ± 0.0002 & 18.37 ± 0.5444 & 86.25 ± 0.1389 & 48.25 ± 0.8187 \\       

LLaMA 1 7B & 2.4 ± 0.0042 & 2.35 ± 0.0027 & 2.28 ± 0.0093 & 17.91 ± 0.1545 & 83.08 ± 0.0483 & 47.57 ± 0.8279 \\
        
LLaMA 1 7B Instruct & 4.7 ± 0.0010 & 4.74 ± 0.0011 & 4.58 ± 0.0017 & 20.3 ± 0.0185 & \textbf{87.38} ± 0.0245 & \textbf{50.28} ± 0.0225 \\
        
LLaMA 2 7B & 1.95 ± 0.0171 & 2.14 ± 0.0101 & 1.8 ± 0.0018 & 15.35 ± 0.5389 & 84.77 ± 0.2790 & 46.21 ± 0.3205\\
        
LLaMA 2 7B Instruct & \textbf{4.86} ± 0.0005 & \textbf{4.9} ± 0.0011& \textbf{4.71} ± 0.0009 & \textbf{22.63} ± 0.1026 & 86.35 ± 0.0327 & 48.34 ± 0.0453 \\
        \hline
    \end{tabular}
    }
    \caption{One-shot performance of the original model, instructed model, and some baseline models on MediQA-Task A: doctor-patient conversation to clinical note and ICliniq: doctor-patient question answering, $p<0.05$; 95\% confidence interval.}
    \label{tab:gen}
\end{table*}

\begin{table}[h!]
\centering
\resizebox{0.8\textwidth}{!}{
\begin{tabular}{|c|ccc|ccc|}
\hline
\multirow{2}{*}{\diagbox{Eval Task}{Train Task}} & \multicolumn{3}{c|}{Performance} & \multicolumn{3}{c|}{Gain} \\
& QA & IE & GEN & w. QA & w. IE & w. GEN \\
\hline
QA &\textbf{53.46} & 52.80 & 52.27 & \textbf{-} & 0.07\% & -0.94\% \\
IE &69.07 & \textbf{72.43} & 65.17 & 2.04\% & - & -3.74\% \\
GEN &4.36 & 4.39 & \textbf{4.42} & 2.15\% & -2.00\% & \textbf{-} \\
\hline
\end{tabular}
}
\caption{Performance on different tasks when trained on a single task and performance gain when model is trained with one additional task besides the evaluation task. For the QA task, average accuracy was calculated using results from four datasets: MedQA-USMLE, MedMCQA, PubmedQA, and BioASQ MCQA. For the IE task, average F1 scores from Medication Status Extraction and Coreference Resolution were used. Lastly, for the Generation task, average scores based on GPT-4 metrics from the Conv2note and Doctor-Patient QA datasets were evaluated.  }
\label{tab:combined_result}
\end{table}

\subsection*{Performance on QA and NLI tasks}

\paragraph{How does instruction tuning perform on QA and NLI tasks?}
\textcolor{black}{Table \ref{tab:medical-model-comparison} details the zero-shot performance across various models on multiple-choice question answering (QA) tasks and natural language inference (NLI) tasks within the biomedical domain, including MedQA-USMLE, MedMCQA, PubmedQA, BioASQ MCQA, and MedNLI. Focusing on the LLaMA series, the instructed variants consistently outperform their non-instructed counterparts, demonstrating the utility of instruction tuning in enhancing model performance. For instance, in the MedQA-USMLE task, instruction-tuned LLaMA 1 7B outperformed its non-instructed variant by a 95\% confidence interval (CI) of 3.96 to 4.10, $p<0.05$. Similar improvements were observed in the MedMCQA, where the instructed version surpassed the non-instructed LLaMA 1 7B by a 95\% CI of 6.2 to 6.6, $p<0.05$. The BioASQ MCQA and PubmedQA tasks further illustrate the strength of instruction tuning. LLaMA 1 13B Instruct exhibited superior performance, with notable improvements in BioASQ MCQA, achieving a score of 84.29 compared to the non-instructed score of 69.86, with a 95\% CI of 11.41 to 17.45, $p<0.05$. Similarly, in PubmedQA, the instructed variant recorded an improvement with a 95\% CI of 2.23 to 5.51, $p<0.05$ over the non-instructed model. Moreover, instruction-tuned LLaMA 2 models also showed remarkable performance leaps. In the MedNLI task, LLaMA 2 13B Instruct achieved an improvement with a 95\% CI of 4.57 to 6.95, $p<0.05$, compared to its non-instructed version. The consistency of improvement across multiple models and tasks validates the effectiveness of instruction tuning in leveraging model capabilities for specific domain tasks. These statistically significant improvements emphasize the critical role of instruction tuning in enhancing the specialized performance of language models, particularly in demanding domains like medical question answering and inference tasks.}

\subsection*{Performance on IE tasks}
\paragraph{How does instruction tuning perform on medication status extraction?} \textcolor{black}{Table \ref{tab:clinical-extraction} showcases the performance of various models on the clinical information extraction task, specifically focused on medication status. Notably, LLaMA 2 7B Instruct recorded the highest F1 score of 75.63 significantly outperforming its non-instructed counterpart by a 95\% CI: 3.40 to 4.42, $p<0.05$. This was paralleled by a substantial boost in precision, achieving the top score of 75.44 compared to 65.58 from its non-instructed version. Similarly, LLaMA 1 7B Instruct showed a notable increase in recall from 71.95 to 84.43, leading to an F1 improvement reflecting with a 95\% CI of 2.10 to 2.68, $p<0.05$. These improvements underline the effectiveness of instruction tuning in enhancing the model's ability to accurately extract and classify medication status, substantiating the method's value in clinical applications.}

\paragraph{How does instruction tuning perform on clinical coreference resolution?} 
\textcolor{black}{Table \ref{tab:clinical-extraction} also illustrates the performance of various models on the clinical information extraction task of coreference resolution. LLaMA 2 7B Instruct led with an F1 score of 61.24, markedly better than its non-instructed version which posted an F1 of 50.03. This improvement is with a 95\% CI: 10.88 to 11.54, $p<0.05$ highlights the substantial enhancement brought about by instruction tuning. The LLaMA 1 7B Instruct also saw a significant F1 increase to 55.14 with a 95\% CI: 10.56 to 11.86, $p<0.05$. The advancements in precision and recall across both models accentuate the advantage of instruction tuning in mastering the intricacies of coreference resolution within the clinical domain, which is essential for constructing coherent and comprehensive patient narratives.}

\subsection*{Performance on Generation tasks}
\paragraph{How does instruction tuning perform on Short Dialogue2Note Summarization?} \textcolor{black}{Table \ref{tab:gen} encapsulates the performances of various models on the Conv2note, a challenging task aimed at converting doctor-patient conversations to clinical notes. The instruction-tuned LLaMA 1 7B model demonstrated significant improvements across all metrics compared to its non-instructed version, with a marked increase in Coherence from 2.74 to 4.49, Completeness from 2.76 to 4.54, and Naturalness from 2.44 to 4.42, outperforming the baseline LLaMA 1 7B by a 95\% confidence intervals of 1.747-1.753, 1.777-1.782, and 1.975-1.985, respectively, all with $p<0.05$. Similarly, the LLaMA 2 7B Instruct model also exhibited notable improvements, achieving higher scores in all categories than the non-instructed version, which highlights the effectiveness of instruction tuning in enhancing the model's ability to generate coherent, complete, and natural notes from conversations.}

\paragraph{How does instruction tuning perform on Doctor-Patient QA?} 
Table \ref{tab:gen}  also illustrates the performances of several models on ICliniq, a challenging task aimed at mimicking the doctor-patient question answering. LLaMA 2 7B Instruct not only surpassed its non-instructed counterpart significantly with improvements in Coherence (1.95 to 4.86), Completeness (2.14 to 4.9), and Naturalness (1.8 to 4.71) but also outperformed the PMC-LLaMA 7B and ChatDoctor models. These improvements are statistically significant with a 95\% confidence intervals of 2.90-2.93, 2.748-2.772, and 2.908-2.912, respectively, all with $p<0.05$., showcasing instruction tuning's pivotal role in optimizing models for complex QA tasks. The LLaMA 2 7B Instruct also set a new benchmark by achieving the highest scores across almost all metrics, further underscoring the impact of targeted instruction tuning in enhancing model performance on specialized tasks.

\begin{figure*}[ht!]
\begin{center}
\includegraphics[width=1.0\textwidth]{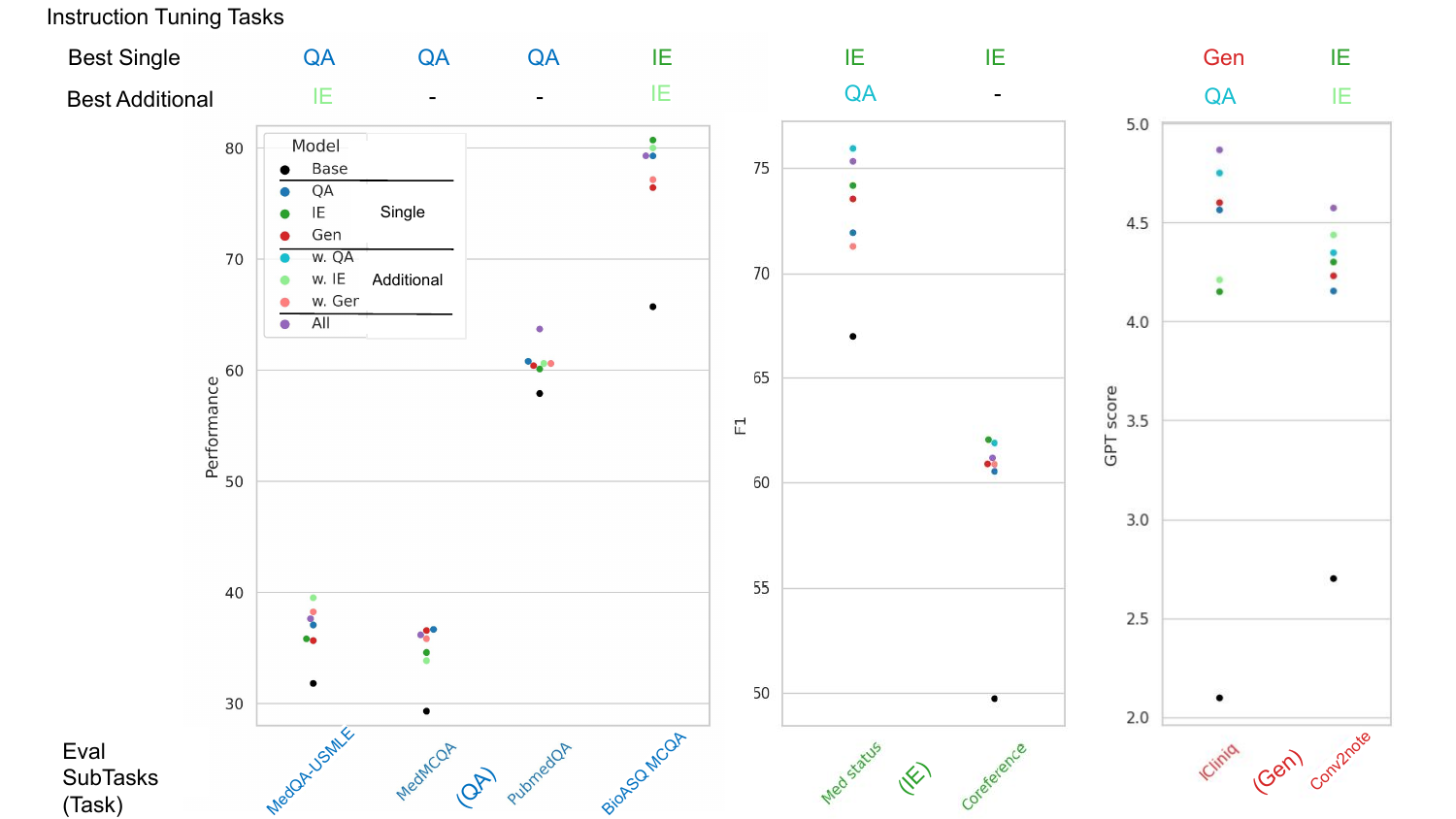}
\caption{Performance of different tasks in BioInstruct. Each scatter corresponds to a subtask to evaluate. Each colored dot inside the scatter represents a different training task. The black dot represents the baseline performance of LLaMA 2 7B without BioInstruct fine-tuning. The purple dot represents the performance of LLaMA 2 7B fine-tuned on all BioInstruct tasks. We then ablate BioInstruct. Above each scatter, we provide the best single task fine-tuned (dark blue, green, red) in the 1st row. In the 2nd row, we also provide the best fine-tuning task in addition to the specific task A, where task A is the same as the evaluation task (light blue, green, red).}

\label{fig:multitask}
\end{center}
\end{figure*}

\subsection*{Multi-task Instruction Tuning Result}

In the previous section, we find that models instruction-fine-tuned with BioInstruct outperformed models without BioInstruct. In this section, we attempt to find out how BioInstruct impact model performance. We summarize our findings by answering the following 4 questions:

\noindent \textbf{Which task-specific tuning results in the best model performance?} Table \ref{tab:combined_result} contains the result of our LLaMA2 model instruction fine-tuned with only one single task. For each evaluation subtask, model instruction fine-tuned on the same task as evaluation task outperforms model instruction fine-tuned on other tasks. Except two non-IE subtasks: BioASQ MCQA and Conv2note. 

\noindent \textbf{Which additional task-specific tuning results improve the most?} We also explore synergies between tasks in multi-task scenario. Table \ref{tab:combined_result} also contains the result of our LLaMA 2 model instruction fine-tuned with two tasks combined. During instruction tuning, we fix one training task the same as the evaluation task and permutate the additional task among tasks other than the evaluation task. We find that models fine-tuned on some additional tasks outperformed those fine-tuned without the additional task, but not all additional tasks improves performance in evaluation task. We summarize our findings as follows:

\begin{enumerate}
[leftmargin=.2in,topsep=2pt]
\setlength\itemsep{0.01em}
\vspace{-0.2em}
    \item Additional IE task helps QA (0.07\%) and vice versa (2.04\%).
    \item The additional QA task benefits the generative task by 2.15\%. However, the generative task does not benefit QA.
\end{enumerate}

\noindent \textbf{Does fine-tuning more tasks achieves better performance?} Finally, we explore if more training tasks leads to better performance. We verify the following statement for different evaluation tasks: Model fine-tuned on all tasks outperforms model fine-tuned on single task or two tasks as mentioned previously. We found that this statement is true for all generative tasks. But it does not apply to any of the IE tasks as shown in Figure \ref{fig:multitask}.

\begin{figure*}[ht!]
\centering

\begin{subfigure}{0.6\textwidth}
\centering
\includegraphics[width=\textwidth]{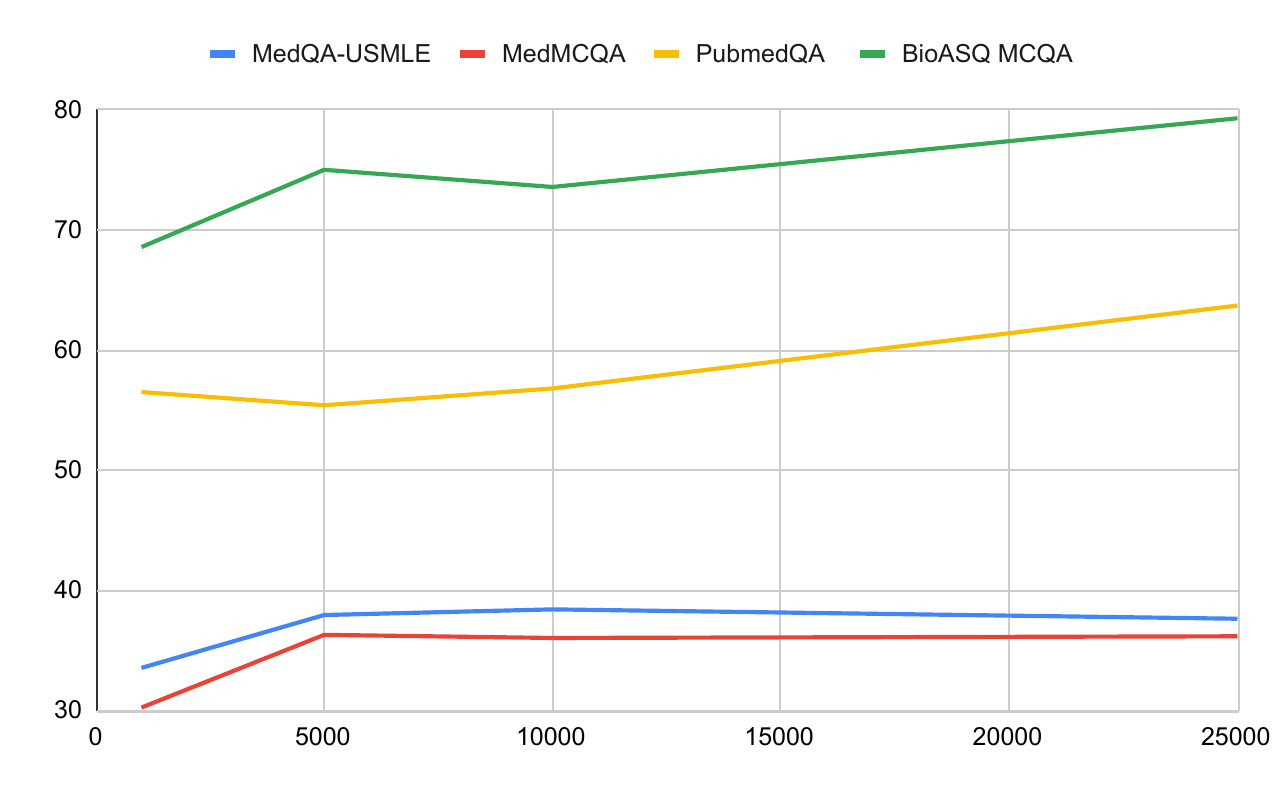}
\end{subfigure}
\hfill
\begin{subfigure}{0.6\textwidth}
\centering
\includegraphics[width=\textwidth]{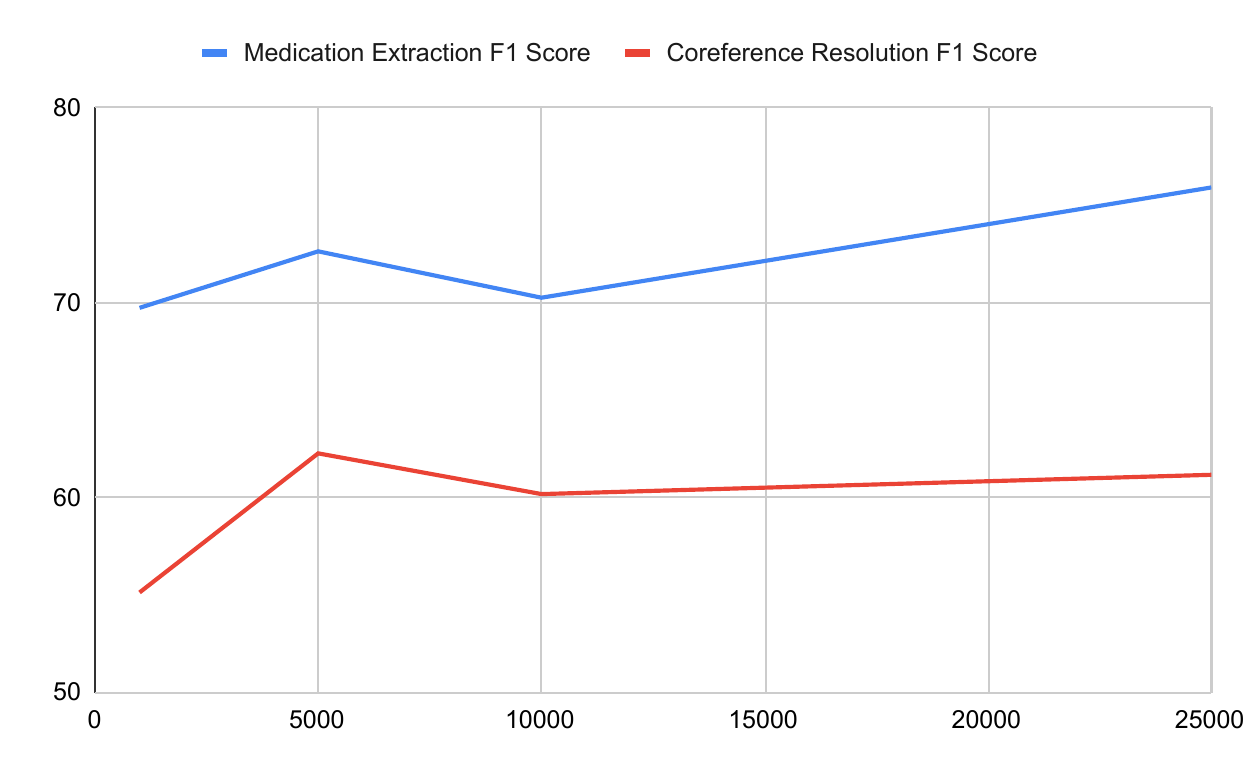}
\end{subfigure}
\hfill
\begin{subfigure}{0.6\textwidth}
\centering
\includegraphics[width=\textwidth]{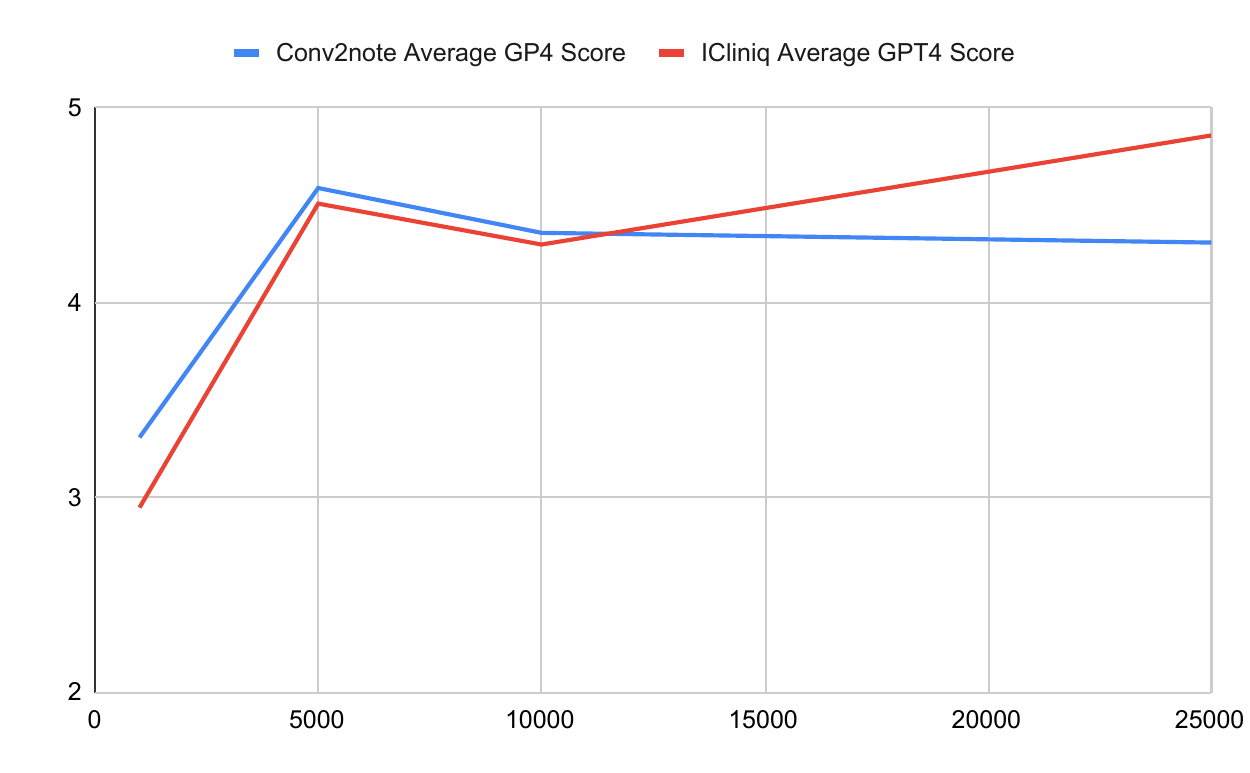}
\end{subfigure}

\caption{Performance on different evaluation tasks when LLaMA 2 7B is fine-tuned on varying number of instruction samples in BioInstruct.}
\label{fig:datasize}
\end{figure*}

\noindent \textbf{How much data used for instruction tunning will make a difference?} We found that 5K samples from BioInstruct significantly improve the performance on most evaluation task as shown in Figure \ref{fig:datasize}. However, adding more fine-tuning samples would still improve the performance. Further investigation into performance dynamics revealed a nuanced response to increasing instruction volumes across different task categories. Specifically, while QA and IE tasks generally benefited from a larger set of instructions, a notable dip in IE task performance at the 10,000 samples suggested a temporary challenge in generalization, which was subsequently overcome as instruction diversity increased at the 25,000 samples. Conversely, generation tasks demonstrated a steady improvement in performance with the addition of more instructions, reinforcing the positive impact of data volume on model capability. This analysis confirms that beyond the initial significant improvements seen with 5,000 samples, further enriching the instruction pool continues to enhance model performance, highlighting the importance of both the quantity and diversity of instructional data.

\section*{Discussion}

In this study, we attempted to find out whether the impact of BioInstruct is tailored to specific BioNLP applications. For this, we draw the synergies from multitask learning \cite{Zhang2018AnOO, Li2019FineTuningBE}, where studies have shown that similar tasks lead to better improvement than tasks that differ (Table \ref{tab:combined_result}). As shown in Figure \ref{fig:multitask}, the highest performance gain resides in the scenario when the task and its intructions are similar. 
In addition, our results show that instruction tuning with additional QA tasks improve performance when evaluating on IE and generative task, compared to those without additional QA tasks.
This supported previous finding in transfer learning: intermediate training on reasoning task tend to transfer the best in various downstream tasks \cite{pruksachatkun-etal-2020-intermediate}.

As shown in Figure \ref{fig:datasize}, instruction tuning with more data improved the performance for most task except two clinical QA tasks: MedQA-USMLE and MedMCQA, especially when compared to biomedical QA tasks: PubmedQA and BioASQ. These tasks primarily focus on information retrieval from biomedical literature, where an extensive dataset provides a richer context and a broader knowledge base for the model to understand and generate accurate responses. On the other hand, MedQA-USMLE and MedMCQA present a set of challenges distinct from the former tasks, as they are centered around clinical questions, which tend to be more specific, scenario-based, and require a depth of understanding in clinical reasoning. The narrower focus and the need for practical application of knowledge in these clinical tasks might not benefit as much from the addition of more data, especially if the added data is more general to biomedical contexts. 

Unfortunately, these was no best instruction tuning task combination that achieves the best performance across all BioNLP tasks (Figure \ref{fig:multitask}). It varied depending on the specific evaluation task. This correlated with previous studies in the general NLP domain \cite{McCann2018TheNL, wang-etal-2019-tell, pruksachatkun-etal-2020-intermediate}, and it was unclear how to identify such subset \cite{Aribandi2022ExT5TE}. Future work could explore task embedding for meta-learning \cite{vu-etal-2020-exploring,Kim2023TaskWebSB} in the BioNLP domain.

\section*{Limitation}

Despite the merits of this study, there are several limitations that could be improved upon in future research. First, 
we adopted an empirical sampling approach from prior research \cite{wang2022self}, choosing 3 demonstrations from a pool of examples to construct the BioInstruct dataset. Future research could benefit from experimenting with the number of demonstrations as a hyperparameter.
Second, our selection of demonstrations was random. Previous works have indicated that choosing demonstrations with semantic similarities can enhance performance in downstream tasks \cite{liu-etal-2022-makes}, whereas others have suggested benefits from a more diverse selection \cite{Su2022SelectiveAM}. An in-depth analysis of the balance between similarity and diversity could provide more insights for future investigations. 
Additionally, a comparison with closed-source models is missing in our current evaluation, which could provide a more comprehensive benchmark of our model's capabilities. This aspect will be considered in subsequent studies to better position our findings within the broader landscape of model performances.
Finally, we found that adding more fine-tuning samples would still improve the performance of some tasks. 
In future work we will explore the threshold for the best performance.

\section*{Conclusion}
We introduce BioInstruct, an automatically generated dataset of natural language instructions and their corresponding inputs and outputs. Our experiments show that LLMs trained on BioInstruct outperformed not only LLMs without BioInstruct, but also other strong LLMs being fine-tuned on extensive amounts domain-specific data. 
Our results also show that the performance gained the most when instruction tuned on instructions with similar BioNLP applications and that the performance gain was the steepest with the first 5k instructions.

\section*{Acknowledgments}
We thank our colleagues and reviewers for their constructive comments and insights that greatly improved the manuscript.

\section*{FUNDING}
Research reported in this study was supported by the National Institute of Nursing Research and the National Institute of Mental Health of the National Institutes of Health under award numbers 1R01NR020868 and R01MH125027, respectively. This work was supported with funding by the National Center on Homelessness among Veterans, U.S. Department of Veterans Affairs Homeless Programs Office. The content is solely the responsibility of the authors and does not necessarily represent NIMH, NINR, NIH, US Department of Veterans Affairs, or the US government.

\section*{AUTHOR CONTRIBUTIONS}
Hieu Tran is the implementer who built the NLP system. He was responsible for building and analyzing various NLP models and methods proposed in this paper and was also one of the writers of the final paper.

Zhichao Yang participated in the formulation, discussion, and writing of the project with his question answering and information extraction experience, providing many valuable views and suggestions from the perspective of NLP or BioNLP research.

Zonghai Yao participated in the formulation, discussion, and writing of the project with his text generation experience, providing many valuable views and suggestions from the perspective of NLP or BioNLP research.

Hong Yu is the planner of the whole project. She was responsible for writing the proposal for the project, planning the direction and progress of the whole project, discussing and suggesting the establishment of the NLP system, and writing the final paper.

\section*{CONFLICT OF INTEREST STATEMENT}
The authors declare no competing interests.

\section*{DATA AVAILABILITY}
The data will be released here: \url{https://github.com/hieutran81/BioInstruct}

\makeatletter
\renewcommand{\@biblabel}[1]{\hfill #1.}
\makeatother

\bibliographystyle{vancouver}
\bibliography{amia}

\newpage
\appendix
\section*{Appendix}
\label{sec:appendix}
\subsection*{Base Models}
We conduct instruction tuning on two base LLMs from the LLaMA family.
\begin{itemize}
    \item \textbf{LLaMA 1} \cite{touvron2023LLaMA} is the most
widely-used open-source language model, it has been trained on a large text corpus with only auto-regressive learning, i.e., no instruction tuning is involved. We report its 7B and 13B models’ performance. 
    \item \textbf{LLaMA 2} \cite{touvron2023LLaMA2} is an updated version of Llama, trained on a new mix of publicly available data with the increasing size of the pretraining corpus by 40\%, doubled the context length of the model, and adopted grouped-query attention \cite{ainslie2023gqa}. We also report its 7B and 13B models’ performance. 
\end{itemize}

\subsection*{Parameter-Efficient Fine-Tuning}
Standard fine-tuning often requires vast amounts of computational resources, as well as high-quality and extensive datasets. However, given the limited size of our instruction data, it is crucial to adopt methods that are more efficient
in terms of computational cost and data requirements. Parameter-efficient tuning methods \cite{li2021prefix, hu2021lora} help achieve this goal by making better use of the available data and minimizing the need for extensive resource allocation.

Specifically, we employ the Low-Rank Adaptation (LoRA) technique \cite{hu2021lora} to fine-tune the LLaMA 1 and LLaMA 2 models. The essence of LoRA lies in its approach to neural networks, which consist of dense layers responsible for matrix multiplications. Typically, these layers' weight matrices are of full-rank. \cite{aghajanyan2020intrinsic} have demonstrated that when adapting to specific tasks, pre-trained language models possess a low “intrinsic dimension”. Building on this insight, LoRA hypothesize that weight updates during this adaptation phase also exhibit a low “intrinsic rank”. For a pre-trained weight matrix $W_0 \in \mathbb{R}^{d \times k}$ its update is constrained by a low-rank decomposition $W_0 + \Delta W = W_0 + BA$, where $B \in \mathbb{R}^{d \times r}$, $A \in \mathbb{R}^{r \times k}$, and rank $r \ll \min(d, k)$. During training, $W_0$ is frozen and does not receive gradient updates whereas $A$ and $B$ are trainable parameters. For $h = W_0 x$, the modified forward pass become:
\[ h = W_0x + \Delta W x = W_0x + BAx \]

With LoRA, we can efficiently fine-tune the LLMs even with limited GPU memory, enabling the fine-tuning of the 7B and 13B versions of LLaMA 1 and LLaMA 2 using just 2 A100 GPUs. Furthermore, it's only necessary to save the model adaptation weights (matrices $B$ and $A$) as checkpoints. These adaptation weights are significantly more compact than the original model weights, simplifying the process of sharing the models.

\subsection*{Experimental Settings}
We tested the models in two settings: zero-shot and one-shot setting:
In zero-shot setting, the model is only given the task instruction without any training examples and then directly evaluated on the test set.
In one-shot setting, the model is provided with one example from the training set, along with the task instruction, before being evaluated on the test set.
This approach allows us to gauge the models' intrinsic capability to generalize from limited or no exposure to examples.

\begin{table*}[t!]
\begin{small}

\begin{tcolorbox}[colback=white,coltext=black]

You are asked to come up with a set of 20 diverse task instructions specifically related to the biomedical or healthcare domain. These task instructions will be given to a GPT model and we will evaluate the GPT model for completing the instructions.\

Here are the requirements:

1. Try to ensure a diverse set of tasks by varying the actions required in each instruction.

2. The language used for the instruction also should be diverse. For example, you should combine questions with imperative instructions.

3. The type of instructions should be diverse. The list should include diverse types of tasks like question answering, classification, summarization, simplification, etc.

4. A GPT language model should be able to complete the instruction. For example, do not ask the assistant to create any visual or audio output.

5. The instructions should be in English.

6. The instructions should be concise and comprehensive. Either an imperative sentence or a question is permitted.

7. You should generate an appropriate input to the instruction. The input field should contain a specific example provided for the instruction. It should involve realistic data and should not contain simple placeholders. The input should provide substantial content to make the instruction challenging.

8. The output should be an appropriate response to the instruction and the input.

List of 20 tasks:

1. Instruction: Given a piece of medical text, rewrite it in a simpler way, suitable for a general audience without losing the core information.

1. Input:
Acute myocardial infarction (AMI), commonly known as a heart attack, occurs when blood flow decreases or stops to a part of the heart, causing damage to the heart muscle.

1. Output:
A heart attack, also known as acute myocardial infarction, happens when the heart does not get enough blood, which can damage the heart muscle.

2. Instruction: Label medications, ignoring allergies. Include dosage, route, frequency, duration, reason, if available.

2. Input:
8. Albuterol 2 puffs every 4-6 hours as needed. HOSPITAL COURSE: This is an 80-year-old female who was hospitalized about 2 months ago for chronic obstructive pulmonary disease exacerbation. At that time she was put on prednisone and antibiotics and seemed to get better. However, she was put on Augmentin ES and continued to have difficulty tasting food and felt that food tasted very salty. She had no appetite and she has continued to lose weight over the last 2 months.

2. Output:-"medication: "Albuterol", dosage: "2 puffs", frequency: "every 4-6 hours", duration: "as needed" -medication: "prednisone", duration: "2 months" -medication: "antibiotics", duration: "2 months" -medication: "Augmentin ES", duration: "2 months"

3. Instruction: 
Given a specific health-related question from a user, provide a concise and accurate response based on general health knowledge. Avoid providing personal medical advice and remind the user to consult with a healthcare professional for personal health concerns.

3. Input:
What can I do to help prevent heart disease?

3. Output:
Regular exercise, a healthy diet, maintaining a healthy weight, not smoking, limiting alcohol, and controlling high blood pressure, high cholesterol and diabetes can help you prevent heart disease. Always consult with a healthcare professional for personal health concerns.

\color{blue}
4. Instruction: Identify the main conclusion from the provided medical report excerpt.

4. Input:
The patient's blood test results showed an elevation in liver enzymes, specifically ALT and AST, which suggests potential liver damage. Additionally, the patient's ultrasound showed a fatty liver.

4. Output:
The patient has signs of liver damage and a fatty liver.

5. Instruction: ...

...........................
\end{tcolorbox}
\end{small}
\caption{ An illustration of our data generation prompt.
Black: The prompt provided to the model. Blue: Model’s generations for the given prompt.}
\label{fig:prompt}
\end{table*}

\begin{table*}
\begin{center}
\resizebox{0.99\textwidth}{!}{
\begin{tabular}{p{18cm}}
\toprule

\textbf{Examples} \\
\midrule
\textbf{Instruction:} Given a detailed health-related query from a consumer, summarize it into a concise question. 

\textbf{Input:} Hi, I have mixed diagnosis of autoimmune connective tissue disease and would like to submit information about this disease. I also need some serious rheumatologist recommended in Argentina or Latin America. From already thank you very much. 

\textbf{Output:} How can I find rheumatologist(s) who specialize in mixed connective tissue disease in Latin America? \\

\midrule

\textbf{Instruction:} Given a section of a medical record and a question related to it, provide the relevant information from the record as an answer.

\textbf{Input:} 08/31/96 ascending aortic root replacement with homograft with omentopexy. The patient continued to be hemodynamically stable making good progress. Physical examination: BMI: 33.4 Obese, high risk. Pulse: 60. resp. rate: 18. Has the patient ever had an abnormal BMI? 

\textbf{Output:} BMI: 33.4 Obese, high risk \\
\midrule

\textbf{Instruction:} Given a pair of sentences, determine if the second sentence is an entailment, contradiction, or neutral with respect to the first one.

\textbf{Input:} Sentence1: Labs were notable for Cr 1.7 (baseline 0.5 per old records) and lactate 2.4.

Sentence2: Patient has elevated Cr.

\textbf{Output:} entailment \\

\midrule

\textbf{Instruction:} Given a medical scenario and a multiple choice question, choose the correct option.

\textbf{Input:} A 23-year-old pregnant woman at 22 weeks gestation presents with burning upon urination. She states it started 1 day ago and has been worsening despite drinking more water and taking cranberry extract. She otherwise feels well and is followed by a doctor for her pregnancy. Her temperature is 97.7°F (36.5°C), blood pressure is 122/77 mmHg, pulse is 80/min, respirations are 19/min, and oxygen saturation is 98\% on room air. Physical exam is notable for an absence of costovertebral angle tenderness and a gravid uterus. Which of the following is the best treatment for this patient? Options: A) Ampicillin B) Ceftriaxone C) Ciprofloxacin D) Doxycycline E) Nitrofurantoin

\textbf{Output:} E) Nitrofurantoin \\

\midrule

\textbf{Instruction:} generating a section summary associated with the short input conversation. 

\textbf{Input:} 

Doctor: What brings you back into the clinic today, miss? 
 
 Patient: I came in for a refill of my blood pressure medicine. 
 
 Doctor: It looks like Doctor Kumar followed up with you last time regarding your hypertension, osteoarthritis, osteoporosis, hypothyroidism, allergic rhinitis and kidney stones. Have you noticed any changes or do you have any concerns regarding these issues? 
 
 Patient: No. 
 
 Doctor: Have you had any fever or chills, cough, congestion, nausea, vomiting, chest pain, chest pressure?
 
 Patient: No. 
 
 Doctor: Great. Also, for our records, how old are you and what race do you identify yourself as?
 
 Patient: I am seventy six years old and identify as a white female.

\textbf{Output:} The patient is a 76-year-old white female who presents to the clinic today originally for hypertension and a med check. She has a history of hypertension, osteoarthritis, osteoporosis, hypothyroidism, allergic rhinitis and kidney stones. Since her last visit she has been followed by Dr. Kumar. Those issues are stable. She has had no fever or chills, cough, congestion, nausea, vomiting, chest pain, chest pressure. \\

\bottomrule
\end{tabular}

}\end{center} 
\caption{Examples of instructions, inputs, and outputs in our seed tasks.}
\label{fig:examples}
\end{table*}

\begin{table*}[t!]
\begin{small}

\begin{tcolorbox}[colback=white,coltext=black]
Below is an instruction that describes a task, paired with an input that provides further context.

Write a response that appropriately completes the request.

\texttt{\#\#\# Instruction:}

Given a short conversation between patient and doctor, write a section summary associated with that conversation.

\texttt{\#\#\# Input:}

Doctor: What brings you back into the clinic today, miss?

Patient: I came in for a refill of my blood pressure medicine.

Doctor: It looks like Doctor Kumar followed up with you last time regarding your hypertension, osteoarthritis, osteoporosis, hypothyroidism, allergic rhinitis and kidney stones. Have you noticed any changes or do you have any concerns regarding these issues?

Patient: No.

Doctor: Have you had any fever or chills, cough, congestion, nausea, vomiting, chest pain, chest pressure?

Patient: No.

Doctor: Great. Also, for our records, how old are you and what race do you identify yourself as?

Patient: I am seventy six years old and identify as a white female.

\texttt{\#\#\# Output:}:

\color{blue}
The patient is a 76-year-old white female who presents to the clinic today originally for hypertension and a med check. She has a history of hypertension, osteoarthritis, osteoporosis, hypothyroidism, allergic rhinitis and kidney stones. Since her last visit she has been followed by Dr. Kumar. Those issues are stable. She has had no fever or chills, cough, congestion, nausea, vomiting, chest pain, chest pressure.
\end{tcolorbox}
\end{small}
\caption{ An illustration of our instruction tuning prompt.
Black: The prompt provided to the model. Blue: Model’s generations for the given prompt.}
\label{fig:instruction-tuning}
\end{table*}

\end{document}